\documentclass[twoside]{article}

\usepackage[accepted]{aistats2024}

\usepackage{hyperref}
\hypersetup{
    colorlinks=true,
    linkcolor=blue,
    citecolor=cyan,
}
\usepackage{xcolor}
\usepackage{amsthm,amssymb,amsmath}
\usepackage{mathrsfs}
\usepackage{mathtools}

\usepackage{enumitem}
\setlist[itemize,0]{itemsep=0pt,leftmargin=0.5cm,topsep=0pt}

\usepackage{algorithm,algpseudocode}

\newtheoremstyle{definition}
{3pt} 
{3pt} 
{} 
{} 
{\bfseries} 
{.} 
{.5em} 
{} 

\DeclareMathOperator*{\argmax}{arg\,max}
\DeclareMathOperator*{\argmin}{arg\,min}

\theoremstyle{definition}
\newtheorem{theorem}{Theorem}
\newtheorem{proposition}[theorem]{Proposition}

\usepackage{xspace}
\usepackage{bbm}
\newcommand{\E}{\mathop{\mathbb{E}}}

\usepackage[round]{natbib}

\newcommand{\piref}{\pi_\text{ref}}
\newcommand{\ipo}{\texttt{IPO}\xspace}

\newcommand{\dpo}{\texttt{DPO}\xspace}

\newcommand{\slic}{\texttt{SLiC-HF}\xspace}
\newcommand{\rlhf}{\texttt{RLHF}\xspace}
\newcommand{\sipo}{$\Psi$\texttt{PO}\xspace}
\usepackage{verbatim}
\newcommand{\1}[1]{\mathbb{I}\{#1\}}

\begin{document}

\twocolumn[

\aistatstitle{A General Theoretical Paradigm to 
Understand Learning from Human Preferences}

\aistatsauthor{Mohammad Gheshlaghi Azar \And Mark Rowland \And Bilal Piot }
\aistatsauthor{Daniel Guo \And Daniele Calandriello \And  Michal Valko \And  R\'emi Munos }
\runningauthor{Gheshlaghi Azar, Rowland,  Piot, Guo, Calandriello, Valko, Munos }
\runningtitle{Understand Learning from Human Preferences}

\aistatsaddress{Google DeepMind} ]

\begin{abstract}
The prevalent deployment of learning from human preferences through reinforcement learning (\rlhf) relies 
on two important approximations: the first assumes that pairwise preferences can be substituted with pointwise rewards. The second assumes that a reward model trained on these pointwise rewards can generalize from collected data to out-of-distribution data sampled by the policy. Recently,  Direct Preference Optimisation (\dpo) has been proposed as an approach that bypasses the second approximation and learn directly a policy from collected data without the reward modelling stage. However, this method still heavily relies on the first approximation.

In this paper we try to gain a deeper theoretical understanding of these practical algorithms. In particular we derive a new general objective 
called \sipo for learning from human preferences that is expressed in terms of pairwise preferences and therefore bypasses both approximations. This new general objective allows us to perform an in-depth analysis of the behavior of \rlhf and \dpo (as special cases of \sipo) and to identify their potential pitfalls.  We then consider another special case for \sipo  by setting $\Psi$ simply to Identity, for which we can derive an efficient optimisation procedure, prove performance guarantees and demonstrate its
empirical superiority to \dpo on some illustrative examples.
\end{abstract}

\section{Introduction}

Learning from human preferences~\citep{christiano2017deep} is a paradigm adopted in the natural language processing literature to better align pretrained~\citep{radford2018improving, ramachandran2016unsupervised} and instruction-tuned~\citep{wei2021finetuned} generative language models to human desiderata. It consists in first collecting large amounts of data where each datum is composed of a context, pairs of continuations of the context, also called generations, and a pairwise human preference that indicates which generation is the best. Then, a policy generating \textit{good} generations given a context is learnt from the collected data. We frame the problem of learning from human preferences as an offline contextual bandit problem~\citep{lu2010contextual}. The goal of this bandit problem is that given a context to choose an action (playing the role of the generation) which is most preferred by a human rater under the constraint that the resulting bandit policy should be close to some known reference policy. The constraint of staying close to a known reference policy can be satisfied e.g., by using KL regularisation~\citep{geist2019theory} and its role is to avoid model drift~\citep{lazaridou2020multi, lu2020countering}.

A prominent approach to tackle the problem of learning from human preferences is through reinforcement learning from human feedback  \citep[\rlhf,][]{ouyang2022training,stiennon2020learning} in which first a reward model is trained in the form of a classifier of preferred and dispreferred actions. Then the bandit policy is trained through RL to maximize this learned reward model while minimizing the distance with the reference policy. Recently \rlhf has been used successfully in solving the problem of aligning generative language models with human preferences~\citep{ouyang2022training}. Furthermore recent works such as direct preference optimisation \citep[\dpo,][]{rafailov2023direct} and \citep[\slic,][]{zhao2023slic} have shown that it is possible to optimize the bandit policy directly from human preferences without learning a reward model. They also  have shown that on a selection of standard language tasks they are competitive with the state of the art \rlhf while they are simpler to implement and require less resources.

Despite this practical success, little is known regarding theoretical foundations of these practical methods.
Notable exceptions, that consider specific special cases, are  \citep{wang2023rlhf,chen2022humanintheloop} and prior 
work on preference-based \citep{robi13,BusaFeketeSzorenyiWengChengHullermeier13} and dueling bandits and RL~\citep{pmlr-v124-novoseller20a,pacchiano2023dueling}.
However, these theoretical works focus on providing theoretical guarantees in terms of regret bounds in the standard bandit setting and they do not deal with the practical setting of \rlhf, \dpo and \slic.

In this work, our focus is on bridging the gap between theory and practice by introducing a simple and general theoretical representation of the practical algorithms for learning from human preferences. In particular, we show that it is possible to characterise the objective functions of \rlhf and \dpo as special cases of a more general objective exclusively expressed in terms of pairwise preferences. We call this objective $\Psi$-preference optimisation (\sipo) objective, where $\Psi$ is an arbitrary non-deceasing mapping. We then analyze this objective function in the special cases of \rlhf and \dpo and investigate its potential pitfalls. Our theoretical investigation of \rlhf and \dpo reveals that in principle they can be both vulnerable to overfitting. This is due to the fact that those methods rely on the strong assumption that pairwise preferences can be substituted with ELo-score (pointwise rewards) via a Bradley-Terry (BT) modelisation~\citep{bradley1952rank}. In particular, this assumption could be problematic when the (sampled) preferences are deterministic or nearly deterministic as it leads to over-fitting to the preference dataset at the expense of ignoring the KL-regularisation term (see Sec.~\ref{subsec:overfitting}). We then present a simple solution to avoid the problem of overfitting, namely by setting $\Psi$ to identity in the \sipo. This method is called Identity-\texttt{PO} (\ipo) and by construction bypasses the BT modelisation assumption for preferences (see Sec.~\ref{sec:ipo}). Finally, we propose a practical solution, via a sampled loss function (see Sec.~\ref{sec:sampled ipo loss}), to optimize this simplified version of \sipo empirically and, we compare its performance with $\dpo$  on simple bandit examples, providing empirical support for our theoretical findings (see Sec.~\ref{sec:illustrative examples} and Sec.~\ref{sec:sampled preferences}).  
 
\section{Notations}
\label{sec:notations}
In the remaining,  we build on the notations of \dpo~\citep{rafailov2023direct}. Given a context $x\in \mathcal X$, where $\mathcal X$ is the finite space of contexts, we assume a finite action space $\mathcal{Y}$. A policy $\pi\in\Delta_\mathcal{Y}^\mathcal{X}$ associates to each context $x\in\mathcal{X}$ a discrete probability distribution $\pi(.|x) \in \Delta_\mathcal{Y}$ where $\Delta_\mathcal{Y}$ is the set of discrete distributions over $\mathcal{Y}$. We denote $\mu\in\Delta_\mathcal{Y}^\mathcal{X}$ the behavior policy. From a given context $x$, let $y, y' \sim \mu (x)$ be two actions generated independently by the reference policy. These are then presented to human raters who express preferences for one of the generations, denoted as $y_w\succ y_l$ where $y_w$ and $y_l$ denote the preferred and dispreferred actions amongst $\{ y, y' \}$ respectively. We then write true human preference $p^*(y \succ y'|x)$ the probability of $y$ being preferred to $y'$ knowing the context $x$.  The probability comes from the randomness of the choice of the human we ask for their preference. So $p^*(y\succ y'|x)=\E_h[\1{h\mbox{ prefers } y \mbox{ to } y'\mbox{ given } x}]$, where the expectation is over humans $h$.
We also introduce the expected preference of a generation $y$ over a distribution $\mu$ knowing $x$, noted $p^*(y \succ \mu|x)$, via the following equation:
\begin{equation*}
    p^*(y \succ \mu|x) = \underset{y' \sim \mu(.|x)}{\mathbb{E}}[p^*(y \succ y'|x)] \, .
\end{equation*}
For any two policy  $\pi,\mu\in \Delta_\mathcal{Y}^\mathcal{X} $  and a context distribution $\rho$ we denote the total preference of policy $\pi$ to $\mu$ as

\begin{equation*}
    p^*_{\rho}(\pi \succ \mu|x) = \underset{\substack{x\sim\rho\\y \sim \pi(.|x)}}{\mathbb{E}}[p^*(y \succ \mu|x)] \, .
\end{equation*}

In practice, we do not observe $p^*$ directly, but samples $I(y, y'|x)$ from a Bernoulli distribution with mean $p^*(y \succ y'|x)$ (i.e., $I(y, y'|x)$ is $1$ with probability $p^*(y \succ y'|x)$ and $0$ otherwise). In particular, we assume we have access to the preferences through a dataset of rated generations  $\mathcal D=(x_i,y_i,y_i')_{i=1}^N=(x_i,y_{w,i}\succ y_{l,i})_{i=i}^N$,
where $N$ is the dataset size. In addition, for a general finite set $\mathcal{S}$, a discrete probability distribution $\eta\in\Delta_\mathcal{S}$ and a real function $f\in\mathbb{R}^\mathcal{S}$, we note the expectation of $f$ under $\eta$ as $\mathbb{E}_{s\sim\eta}[f(s)]=\sum_{s\in\mathcal{S}}f(s)\eta(s)$. For a finite dataset $\mathcal D=(s_i)_{i=1}^N$, with $s_i \in \mathcal{S}$ for each $i$, and a real function $f\in\mathbb{R}^\mathcal{S}$, we denote the \emph{empirical expectation} of $f$ under $\mathcal{D}$ as $\mathbb{E}_{s\sim D}[f(s)] = \frac{1}{N}\sum_{i=1}^{N}f(s_i)$.

\section{Background}

\subsection{Reinforcement Learning from Human Feedback (\rlhf)}

The standard \rlhf paradigm \citep{christiano2017deep,stiennon2020learning} consists of two main stages: (i) learning the reward model; (ii) policy optimisation using the learned reward. Here we provide a recap of these stages.

\subsubsection{Learning the Reward Model}
Learning a reward model consists in training a binary classifier to discriminate between the preferred and dis-preferred actions using a logistic regression loss. For the classifier, a popular choice is Bradley-Terry model: for a given context $x$ and action $y$, we denote the pointwise reward, which can also be interpreted as an Elo score, of $y$ given $x$ by $r(x,y)$. The Bradley-Terry model represents the preference function $p(y \succ y'|x)$ (classifier) as a sigmoid of the difference of rewards:
\begin{equation}\label{eq:bt-model}
    p(y \succ y'|x)=\sigma\big(r(x,y)-r(x,y')\big) \, ,
\end{equation}

where $\sigma(\cdot)$ denotes the sigmoid function and plays the role of normalisation. Given the dataset  $\mathcal D=(x_i,y_{w,i}\succ y_{l,i})_{i=1}^N$ one can learn the reward function by optimizing the following logistic regression loss
\begin{equation}\label{eq:bt-obj}
\mathcal L(r) = -\mathbb E_{(x,y_w,y_l)\sim \mathcal D} \bigg[\log\left(p(y_w \succ y_l|x)\right)\bigg] \, .
\end{equation}
Assuming that $p^*(y \succ y'|x)$ conforms to the Bradley-Terry model, one can show that as the size of the dataset $\mathcal D$ grows, $p(y \succ y'|x)$ becomes a more and more accurate estimate of true $p^*(y \succ y'|x)$ and in the limit converges to $p^*(y \succ y'|x)$.

\subsubsection{Policy Optimisation with the Learned Reward}

Using the reward (Elo-score) $r(x,y)$ the \rlhf objective is simply to optimize for the policy $\pi\in\Delta^{\mathcal{X}}_{\mathcal{Y}}$ that maximizes the expected reward  while minimizing the distance between $\pi$ and some  reference policy $\piref\in\Delta^{\mathcal{X}}_{\mathcal{Y}}$ through the following KL-regularized objective function:
\begin{equation}\label{eq:rlhf-obj}
J(\pi)= \mathbb E_{\pi}[r(x,y)] - \tau D_{\text{KL}} (\pi \; || \; \piref) \, ,
\end{equation}
in which the context $x$ is drawn from $\rho$ and the action $y$ is drawn from $\pi(.|x)$. The divergence $D_{\text{KL}}(\pi||\piref)$ is defined as follows:
\begin{equation*}
D_{\text{KL}} (\pi \; || \; \piref) = \mathbb E_{x\sim \rho}[ \text{KL}(\pi(.|x) \; || \; \piref(.|x)) ] \, .
\end{equation*}
where:
\begin{equation*}
\text{KL}(\pi(.|x) \; || \; \piref(.|x)) =  \mathbb E_{y\sim \pi(.|x)}\left[\log\left(\frac{\pi(y|x)}{\piref(y|x)}\right)\right].   
\end{equation*}
The objective in Equation~\eqref{eq:rlhf-obj} is essentially optimized by PPO \citep{schulman2017proximal} or similar approaches.

The combination of \rlhf+PPO has been used with great success in practice  \citep[e.g., InsturctGPT and  GPT-4][]{ouyang2022training,openai2023gpt4}.

\subsection{Direct Preference Optimisation}

An alternative approach to the RL paradigm described above is direct preference optimisation \citep[\dpo; ][]{rafailov2023direct}, which avoids the training of a reward model altogether. The loss that \dpo optimises, given an empirical dataset $\mathcal{D}$, as a function of $\pi$, is given by
\begin{align}\label{eq:dpo-obj}
   \min_{\pi}  \mathbb{E}_{(x, y_w, y_l) \sim \mathcal{D}} \nonumber
    \Bigg[ 
        -\log \sigma\Bigg(
        & \tau \log \left( \frac{\pi(y_w|x) }{\pi(y_l|x)} \right) - \\
           & \tau \log \left( \frac{ \piref(y_w|x) }{\piref(y_l|x) } \right) 
            \Bigg)
        \Bigg] \, .
\end{align}
In its population form, the loss takes on the form
\begin{align}\label{eq:dpo-pop-obj}
    \min_{\pi} \E_{\substack{x \sim \rho \\ y,y' \sim \mu}} \nonumber
    \Bigg[ -p^*(y \succ y' |x ) &
        \log  \sigma\Bigg(
         \tau \log \left( \frac{\pi(y|x) }{\pi(y'|x)} \right) - \\
           & \tau \log \left( \frac{ \piref(y|x) }{\piref(y'|x) } \right) 
            \Bigg)
        \Bigg] \, .
\end{align}

\citet{rafailov2023direct} show that when (i) the Bradley-Terry model in Equation~\eqref{eq:bt-model} perfectly fits the preference data and (ii) the optimal reward function $r$ is obtained from the loss in Equation~\eqref{eq:bt-obj}, then the global optimisers of the \rlhf objective in Equation~\eqref{eq:rlhf-obj} and the \dpo objective in Equation~\eqref{eq:dpo-pop-obj} perfectly coincide. In fact, this correspondence is true more generally; see Proposition~\ref{prop:dpo-rlhf} in Appendix~\ref{sec:additional-results}.

\section{A General Objective for Preference Optimisation}

A central conceptual contribution of the paper is to propose a general objective for \rlhf, based on maximizing a non-linear function of preferences. To this end, we consider a general non-decreasing function $\Psi : [0,1] \rightarrow \mathbb{R}$, a reference policy $\piref\in\Delta^{\mathcal{X}}_{\mathcal{Y}}$, and a real positive regularisation parameter $\tau\in\mathbb{R}^*_+$, and define the \emph{$\Psi$-preference optimisation objective} (\sipo) as
\begin{align}\label{eq:main-obj}
    \max_{\pi} \E_{\substack{x \sim \rho \\ y \sim \pi(.|x) \\ y' \sim \mu(.|x)}}[\Psi(p^*(y \succ y' | x))] - \tau D_{\text{KL}}(\pi \; || \; \piref) \, .
\end{align}
This objective balances the maximisation of a potentially non-linear function of preference probabilities with the KL regularisation term which encourages policies to be close to the reference $\piref$. This is motivated by the form of Equation~\eqref{eq:rlhf-obj}, and we will see in the next subsection that it strictly generalises  both \rlhf and  \dpo, when the BT model holds.

\subsection{A Deeper Analysis of \dpo and \rlhf}

In the remaining, we omit the dependency on $x$ for the ease of notations. This is without losing generality and all the following results are true for all $x\in\texttt{Supp}(\rho)$.

We first connect \dpo and \rlhf with the $\Psi$-preference objective in Equation~\eqref{eq:main-obj}, under the special choice of $\Psi(q) = \log(q / (1-q))$. More precisely, the following proposition establishes this connection.

\begin{proposition}
    Suppose $\Psi(q) = \log(q / (1-q))$. When the Bradley-Terry model holds for $p^*$, that is, there exists $r : \mathcal{Y} \rightarrow \mathbb{R}$ such that
    \begin{align*}
        p^*(y \succ y') = \sigma(r(y)-r(y')) \, ,
    \end{align*}
    then the optimal policy for Equation~\eqref{eq:main-obj}, for the \rlhf objective in Equation~\eqref{eq:rlhf-obj}, and for the standard \dpo objective in Equation~\eqref{eq:dpo-pop-obj} are identical.
\end{proposition}
\begin{proof}
    Note that under the assumption that the Bradley-Terry model holds, we have
    \begin{align*}
        \E_{y' \sim \mu}[\Psi(p^*(y \succ y'))] & = \E_{y' \sim \mu}\left[\Psi\left( \frac{e^{r( y)}}{e^{r( y)} + e^{r( y')}} \right)\right] \\
        & = \E_{y' \sim \mu}[\log(e^{r(y)} / e^{r( y')})] \\
        & = \E_{y' \sim \mu}[ r(y) - r(y') ] \\
        & = r(y) - \E_{y' \sim \mu}[ r( y') ] \, .
    \end{align*}
    This is equal to the reward in Equation~\eqref{eq:rlhf-obj}, up to an additive constant, and so
    it therefore follows that the optimal policy for Equation~\eqref{eq:main-obj} and for optimizing the objective in Equation~\eqref{eq:rlhf-obj} are identical. Further, as shown by \citet{rafailov2023direct}, the optimal policy for the \dpo objective in Equation~\eqref{eq:dpo-pop-obj} and the objective in Equation~\eqref{eq:rlhf-obj} are identical, which gives the statement of the proposition.
\end{proof}

Applying this proposition to the objective function of Equation~\eqref{eq:main-obj}, for which there exists an analytical solution, reveals that under the BT assumption the closed-form solution to \dpo and \rlhf  can be written as
\begin{align}
\label{eq:dpo-opt}
    \pi^*(y) \propto \piref(y)\exp \Big(\tau^{-1} \mathbb{E}_{y' \sim \mu}[\Psi(p^*(y \succ y'))] \Big) \, .
\end{align}
The derivations leading to Equation~\ref{eq:dpo-opt} is a well known result and is provided in App.\,\ref{appendices: argmaximum} for completeness.

\subsection{Weak Regularisation and Overfitting}
\label{subsec:overfitting}

It is worth taking a step back, and asking what kinds of policies the above objective leads us to discover. This highly non-linear transformation of the preference probabilities means that small increases in preference probabilities already close to 1 are just as incentivized as larger increases in preference probabilities around $50\%$, which may be undesirable. The maximisation of logit-preferences, or Elo score in game-theoretic terminology, can also have counter-intuitive effects, even in transitive settings \citep{bertrand2023limitations}.

Consider the simple example where we have two actions $y$ and $y'$ such that $p^* (y \succ y') = 1$, i.e., $y$ is always preferred to $y'$. Then the Bradley-Terry model would require that $(r(y)- r( y')) \rightarrow +\infty$ to satisfy \eqref{eq:bt-model}. If we plug this into the optimal policy \eqref{eq:dpo-opt} then we would get that $\frac{\pi^*(y')}{\pi^*(y)} =0$ (i.e., $\pi^*(y')=0$) irrespective of what constant $\tau$ is used for the KL-regularisation. Thus the strength of the KL-regularisation becomes weaker and weaker the more deterministic the preferences.

The weakness of the KL-regularisation becomes even more pronounced in the finite data regime, where we only have access to a sample estimate of the preference $\hat{p}(y \succ y' )$. Even if the true preference is, e.g., $p^* (y \succ y') = 0.8$, empirically it can be very possible when we only have a few data points to estimate $\hat{p}(y \succ y' ) = 1$, in which case the empirical optimal policy would make $\pi(y')=0$ for any $\tau$. This means that overfitting can be a substantial empirical issue, especially when the context and action spaces are extremely large as it is for large language models.

\emph{Why may standard \rlhf be more robust to this problem in practice?} While a purported advantage of \dpo is that it avoids the need to fit a reward function, we observe that in practice when empirical preference probabilities are in the set $\{0,1\}$, the reward function ends up being \emph{underfit}. The optimal rewards in the presence of $\{0,1\}$ preference probabilities are infinite, but these values are avoided, and indeed regularisation of the reward function has been observed to be an important aspect of \rlhf training in practice \citep{christiano2017deep}. This underfitting of the reward function is thus crucial in obtaining a final policy that is sufficiently regularised towards the reference policy $\piref$, and \dpo, in avoiding the training of the reward function, loses the regularisation of the policy that the underfitted reward function affords.

While standard empirical practices such as early-stopping can still be used as an additional form of regularisation to curtail this kind of overfitting, in the next section, we will introduce a modification of the \sipo objective such that the optimal empirical policy can be close to $\piref$ even when preferences are deterministic.

\section{\ipo: \sipo with identity mapping}
\label{sec:ipo}

We have observed in the previous section that \dpo is prone to overfitting, and this stems from a combination of the unboundedness of $\Psi$, together with not training an explicit reward function. Not training a reward function directly is a clear advantage of \dpo, but we would like to avoid the problems of overfitting as well.

This analysis of \dpo motivates choices of $\Psi$ which are bounded, ensuring that the KL regularisation in Equation~\ref{eq:main-obj} remains effective even in the regime of $\{0,1\}$-valued preferences, as it is often the case when working with empirical datasets. A particularly natural form of objective to consider is given by taking $\Psi$ to be the identity mapping in Equation~\eqref{eq:main-obj}, leading to direct regularized optimisation of \emph{total preferences}:
\begin{align}\label{eq:ipo-obj}
    \max_{\pi} p^*_{\rho}(\pi \succ \mu) - \tau D_{\text{KL}}(\pi \; || \; \piref) \, .
\end{align}

The standard approach to optimize the objective function of Equation~\eqref{eq:ipo-obj} is through \rlhf with the choice of reward $r(y)=p^*(y\succ\mu)$. However both using RL and estimating the reward model $r(y)$ can be costly. Inspired by \dpo one would like to devise an empirical solution for the optimisation problem of Equation~\eqref{eq:ipo-obj} which can directly learn from the preference dataset. Thus it would be able to avoid RL and reward modeling altogether.

\subsection{Derivations and Computationally Efficient Algorithm}

As with \dpo, it will be beneficial to re-express Equation~\eqref{eq:ipo-obj} as an offline learning objective. To derive such an expression, we begin by following the derivation of \citet{rafailov2023direct}, manipulating the analytic expression for the optimal policy into a system of root-finding problems. As in the previous section, we drop dependence on the context $x$ from our notation, as all arguments can be applied on a per-context basis.

\textbf{Root-finding problems.} 
Let $g(y) = \mathbb{E}_{y' \sim \mu}[\Psi(p^*(y \succ y'))]$. Then we have
\begin{align}\label{eq:pi-opt}
    \pi^*(y) \propto \piref(y) \exp(\tau^{-1} g(y)) \, .
\end{align}
For any $y, y' \in \texttt{Supp}(\piref)$, we therefore have
\begin{align}\label{eq:w1}
    \frac{\pi^*(y)}{\pi^*(y')} = \frac{\piref(y)}{\piref(y')}\exp(\tau^{-1}(g(y) - g(y'))) \, .
\end{align}
By letting
\begin{align*}
    h^*(y, y') = \log\left( \frac{\pi^*(y)\piref(y')}{\pi^*(y')\piref(y)} \right)
\end{align*}
and rearranging Equation~\eqref{eq:w1}, we obtain
\begin{align}\label{eq:w2}
    h^*(y, y') = \tau^{-1} \big( g(y) - g(y') \big) \, .
\end{align}
The core idea now is to consider a policy $\pi$, define
\begin{align*}
    h_\pi(y, y') = \log\left(\frac{\pi(y) \piref(y'}{\pi(y') \piref(y)}\right) \, ,    
\end{align*}
and aim to solve the equations:
\begin{align}\label{eq:fp}
    h_\pi(y, y') = \tau^{-1} \big( g(y) - g(y') \big) \, .
\end{align}

\textbf{Loss for \ipo.} We now depart from the approach to the analysis employed by \citet{rafailov2023direct}, to obtain a novel offline formulation of Equation~\eqref{eq:main-obj}, in the specific case of $\Psi$ as the identity function.
In this case, Equation~\eqref{eq:fp} reduces to
\begin{align*}
    h_\pi(y, y') = \tau^{-1} \big( p^*(y  \succ \mu) - p^*(y ' \succ \mu)\big) \, .
\end{align*}
We begin by re-expressing these root-finding problems as a single optimisation problem $L(\pi)$:
\begin{align}\label{eq:ipo-loss}
   L(\pi)= \E_{y, y' \sim \mu}\left[ \left( h_\pi(y, y') - \frac{p^*(y \succ \mu) - p^*(y' \succ \mu)}{\tau} \right)^2  \right] \, .
\end{align}

One can easily show that for the choice of $\pi^*$ we have $L(\pi^*)=0$. Thus $\pi^*$ is a global minimizer of $L(\pi)$. The following theorem establishes the uniqueness of this solution.

\begin{theorem}[Uniqueness of Global/Local Optima]
\label{th:root finding}
Assume that $\texttt{Supp}(\mu)= \texttt{Supp}(\piref)$ and define $\Pi$ to be the set of policies $\pi$ such that $\texttt{Supp}(\pi)= \texttt{Supp}(\mu)$. Then $\pi\mapsto L(\pi)$ has a unique local/global minimum in $\Pi$, which is $\pi^*$. 
\end{theorem}

\begin{proof}
By assumption, $\pi^* \in \Pi$, and by definition $\forall \pi\in\Pi, L(\pi)\geq0$ as $L(\pi)$ is an expectation of squared terms.  Further, from Equation~\eqref{eq:w2}, it follows immediately that $L(\pi^*) = 0$, and so we deduce that $\pi^*$ is a global optimum for $L$. We now show that there are no other local/global minima for $L$ in $\Pi$.

We write $J = \texttt{Supp}(\mu)$. We parametrise the set $\Pi$ via vectors of logits $s \in \mathbb{R}^J$, setting $\pi_s(y)=\exp(s(y))/\sum_{y' \in J}\exp(s(y'))$ for $y \in J$, and $\pi_s(y) = 0$ otherwise.   Let us write ${\cal L}(s)=L(\pi_s)$ for the objective as a function of the logits $s$.
\begin{align}
    \label{eq:opt.s.obj}
    {\cal L}(s)&=\mathbb E_{y,y'\sim \mu}\bigg[\bigg[ \frac{p^*(y\succ\mu) - p^*(y'\succ\mu)}{\tau}
    \\&\notag
    -(s(y)-s(y'))-\log\left(\frac{\piref(y')}{\piref(y)}\right)\bigg]^2\bigg] \, .
\end{align}

The objective is quadratic as a function of the logits $s$. Further, by expanding the quadratic above, we see that the loss can be expressed as a sum of squares
\begin{align}\label{eq:quadratic}
    \sum_{y,y' \in J} \mu(y) \mu(y') (s(y) - s(y'))^2
\end{align}
plus linear and constant terms. This is therefore a positive-semidefinite quadratic, and hence is convex. We thus deduce that all local minimisers of the loss $\mathcal L(s)$ are global minimisers as well ~\citep[Chap. 4]{boyd2004convex}.
We now notice since $\pi_s$ is a surjective continuous mapping from $s$ to $\pi$ one can easily show from the definition of local minimum that every local minimiser $\pi$ of $L$ corresponds to a set of local minimisers $\mathcal S_{\pi}$  of $\mathcal L$. Thus all local minimums of $L$ are also global minimums as well.

Finally, the only direction $s$ in which the quadratic in Equation~\eqref{eq:quadratic} does not increase away from 0 is when all bracketed terms remain 0; that is, in the direction $(1,\ldots,1) \in \mathbb{R}^J$. Thus, $\mathcal{L}(s)$ is strictly convex, except in the direction $(1,\ldots,1)$.~\citep[Chap. 3]{boyd2004convex}. However, modifying logits in the direction $e=(1,\ldots,1)$ does not modify the resulting policy $\pi_s$, since, for $y \in J$,
\begin{align*}
    \pi_{s + \lambda e}(y) = \frac{e^{s(y) + \lambda}}{\sum_{y' \in J} e^{s(y') + \lambda}} 
    = \frac{e^{s(y)}}{\sum_{y' \in J} e^{s(y')}}
    = \pi_s(y) \, .
\end{align*}
The strict convexity combined with the fact that $\pi^*$ is a global minima proves that $\pi^*$ is the unique global/local minima in $\Pi$~\citep[Chap. 4]{boyd2004convex}. 
\end{proof}

\subsection{Sampled Loss for \ipo}
\label{sec:sampled ipo loss}

In order to obtain the sampled loss for \ipo we need to show that we can build  an unbiased estimate of the right-hand side of the equation~\eqref{eq:ipo-loss}. To this end, we consider the {\bf Population  \ipo Loss}:
\begin{align}\label{eq:ipo-loss-sa}
    \E_{y, y' \sim \mu}\Big\lbrack \big( h_\pi(y, y') - \tau^{-1}  I(y, y') \big)^2  \Big\rbrack \, ,
\end{align}
where $I(y, y')$ is drawn from a Bernoulli distribution with mean $p^*(y \succ y')$, i.e., $I(y, y')$ is $1$ if $y$ is preferred to $y'$ (which happens with probability $p^*(y \succ y')$), and $0$ otherwise. This straightforwardly yields a sample-based loss that can be used, by sampling a pair $(y, y')$ from the preference dataset, and consulting the recorded preference to obtain a sample from $I(y, y')$. The following proposition justifies the switch from Equation~\eqref{eq:ipo-loss} to Equation~\eqref{eq:ipo-loss-sa}, by demonstrating their equality.
\begin{proposition}
    The expressions in Equation~\eqref{eq:ipo-loss} and Equation~\eqref{eq:ipo-loss-sa} are equal, up to an additive constant independent of $\pi$.
\end{proposition}
\begin{proof}
    This equivalence is not completely trivial, since in general the conditional expectation
    \begin{align*}
        \mathbb{E}[h_\pi(Y, Y') - \tau^{-1} I(Y, Y') \; | \; Y=y, Y'=y']
    \end{align*}
    is not equal to the corresponding quantity appearing in Equation~\eqref{eq:ipo-loss}, namely
    \begin{align*}
        h_\pi(y, y') - \tau^{-1} \big( p^*(y \succ \mu) - p^*(y' \succ \mu)\big) \, .
    \end{align*}
    We instead need to exploit some symmetry between the distributions of $y$ and $y'$, and use the fact that $h_\pi(y, y')$ decomposes as an additive function of $y$ and $y'$. To show this equality of losses, it is enough to focus on the ``cross-terms'' obtained when expanding the quadratics in Equations~\eqref{eq:ipo-loss} and \eqref{eq:ipo-loss-sa}; that is, to show
    \begin{align*}
        & \E_{y, y' \sim \mu}\Big\lbrack h_\pi(y, y') I(y, y') \Big\rbrack\\
        =& \E_{y, y' \sim \mu}\Big\lbrack h_\pi(y, y') (p^*(y \succ \mu) - p^*(y' \succ \mu)) \Big\rbrack \, .
    \end{align*}
    Now, starting with the right-hand side, and using the shorthand $\pi_y = \log(\pi(y))$, $\pi^\text{R}_y = \log(\piref(y))$, $p_y = p^*(y \succ \mu)$, and similarly for $y'$, we have
    \begin{align*}
        & \E_{y, y' \sim \mu}\Big\lbrack h_\pi(y, y') (p^*(y \succ \mu) - p^*(y' \succ \mu)) \Big\rbrack \\
        = & \E_{y, y' \sim \mu}\Big\lbrack (\pi_y - \pi_{y'} + \pi^\text{R}_{y'} - \pi^{\text{R}}_y) (p_y - p_{y'}) \Big\rbrack \\
        = & \E_{y, y' \sim \mu}\Big\lbrack \pi_y p_y - \pi_y p_{y'} - \pi_{y'} p_y + \pi_{y'}\\
        & + p_{y'} + \pi^{\text{R}}_{y'} p_y - \pi^{\text{R}}_{y'} p_{y'} - \pi^{\text{R}}_y p_y + \pi^{\text{R}}_y p_{y'} \Big\rbrack \\
        = & \E_{y, y' \sim \mu}\Big\lbrack (2 p_y - 1) \pi_y - (2 p_y - 1) \pi^{\text{R}}_y \Big\rbrack \, ,
    \end{align*}
    where we have used iid-ness of $y$ and $y'$, and $\mathbb{E}_{y\sim \mu}[p_y]=1/2$. Turning to the left-hand side, we have
    \begin{align*}
        & \E_{y, y' \sim \mu}\Big\lbrack h_\pi(y, y') I(y, y') \Big\rbrack \\
        = & \E_{y, y' \sim \mu}\Big\lbrack \big( \pi_y - \pi_{y'} + \pi^{\text{R}}_{y'} - \pi^{\text{R}}_y \big) I(y, y') \Big\rbrack \\
        = & \E_{y\sim \mu}\Big\lbrack \big( \pi_y - \pi^{\text{R}}_y \big) \E_{y'\sim \mu}[I(y, y') \mid y]\Big\rbrack \\
         & + \E_{y'\sim \mu}\Big\lbrack \big( -\pi_{y'} + \pi^{\text{R}}_{y'} \big) \E_{y\sim \mu}[I(y, y') \mid y' ]\Big\rbrack \\
        = & \E_{y, y' \sim \mu}\Big\lbrack \pi_y p_y - \pi_{y'} (1- p_{y'}) + \pi^{\text{R}}_{y'} (1-p_{y'}) - \pi^{\text{R}}_y p_y \Big\rbrack \\
        = & \E_{y, y' \sim \mu}\Big\lbrack (2 p_y - 1)\pi_y - (2 p_{y} - 1)\pi^R_{y}  \Big\rbrack \, ,
    \end{align*}
    where we use the fact that $\E_{y'\sim \mu}I(y, y')=p_{y}$ and $\E_{y\sim \mu}I(y, y')=1-p_{y'}$.
    This demonstrates equality of the losses, as required.
\end{proof}

We now discuss how to approximate the loss in Equation~\eqref{eq:ipo-loss-sa} with an empirical dataset. As in our earlier discussion, the empirical dataset $\mathcal{D}$ takes the form $(y_{w,i}, y_{l,i})_{i=i}^N$. Note that each datapoint $( y_{w_,i}, y_{l, i})$ contributes two terms to an empirical approximation of Equation~\eqref{eq:ipo-loss-sa}, with $(y, y', I(y, y')) = (y_{w,i}, y_{l,i}, 1)$, and also $(y, y', I(y, y')) = (y_{l,i}, y_{w,i}, 0)$. This symmetry is important to exploit, and leads to a reduction in the variance of the loss. The overall empirical loss is therefore given by
\begin{align*}
    \frac{1}{2}\underset{(y_w,y_l)\sim D}{\mathbb E} \Big\lbrack  (h_\pi(y_{w}, y_{l}) - \tau^{-1})^2 +  h_\pi(y_{l}, y_{w} )^2  \Big\rbrack &=  
    \\
    \frac{1}{2}\underset{(y_w,y_l)\sim D}{\mathbb E} \Big\lbrack  (h_\pi(y_{w}, y_{l}) - \tau^{-1})^2 +  h_\pi(y_{w}, y_{l} )^2  \Big\rbrack &,
\end{align*}

which  up to a constant equals:

\begin{align}
\label{eq:sampled IPO loss}
     \underset{(y_w,y_l)\sim D}{\mathbb E}\left[ \left(h_\pi(y_{w}, y_{l}) - \frac{\tau^{-1}}2\right)^2\right] .
\end{align}

This simplified form of the loss  provides some valuable insights on the way in which \ipo optimizes the policy $\pi$:   \ipo learns from preferences dataset simply  by regressing the gap between log-likelihood ratios $\log(\pi(y_{w})/\pi(y_{l}))$ and $\log(\piref(y_{w})/\piref(y_{l}))$  to $\frac{\tau^{-1}}2$. So the weaker the regularisation becomes, the higher would be the log-likelihood ratio of $y_w$ to $y_l$. In other words  \ipo , unlike \dpo, always  regularizes its solution towards $\piref$ by controlling the gap between the log-likelihood ratios $\log(\pi(y_{w})/\pi(y_{l}))$ and $\log(\piref(y_{w})/\piref(y_{l}))$, thus avoiding the over-fitting to the preference dataset. We summarize the sampled  \ipo in Algorithm \ref{algo:sample.ipo}:

\begin{algorithm}
\caption{Sampled \ipo}
\label{algo:sample.ipo}
\begin{algorithmic}[1]
\Require Dataset $\mathcal D$ of prompts, preferred and dis-preferred generations $x$, $y_w$ and $y_l$, respectively. A reference policy $\piref$
\State Define
\begin{align*}
    h_\pi(y, y',x) = \log\left( \frac{\pi(y|x)\piref(y'|x)}{\pi(y'|x)\piref(y|x)} \right)
\end{align*}

\State Starting from $\pi=\piref$ minimize
\begin{align*}
     \underset{(y_w,y_l,x)\sim D}{\mathbb E} \left(h_\pi(y_{w}, y_{l},x) - \frac{\tau^{-1}}2\right)^2 .
\end{align*}

\end{algorithmic}
\end{algorithm}

\subsection{Illustrative Examples}
\label{sec:illustrative examples}
To illustrate the qualitative difference between our algorithm and \dpo we will consider a few simple cases. For simplicity we assume there is no context $x$, i.e., we are in the bandit setting.

\subsubsection{Asymptotic Setting}
We first consider the simple case where we have 2 actions only, $y_1$ and $y_2$, and a deterministic preference between them: $p^*(y_1 \succ y_2) = 1$. 
Suppose we start with a uniform $\piref$ and $\mu$. We know from Section \ref{subsec:overfitting} that \dpo will converge to the deterministic policy $\pi^*(y_1)=1$, $\pi^*(y_2) = 0$ regardless of the value of $\tau$. Thus even when the regularisation coefficient $\tau$ is very large, this is very different from the uniform $\piref$.

Now, let us derive the optimal policy for IPO. We have $p^*(y_1\succ \mu)= 3/4$ and $p^*(y_2\succ \mu)= 1/4$. Plugging this into equation \eqref{eq:pi-opt} with $\Psi = I$ we get that $\pi^*(y_1) = \frac{\exp(0.75\tau^{-1}) }{\exp(0.75\tau^{-1}) + \exp(0.25\tau^{-1}) } = \sigma(0.5\tau^{-1})$, and $\pi^*(y_2) =  \sigma(-0.5\tau^{-1})$, where $\sigma$ is the sigmoid function. Hence we see that if we have large regularisation as $\tau \rightarrow +\infty$, then $\pi^*$ converges to the uniform policy $\piref$, and on the flip side as $\tau \rightarrow +0$, then $\pi^*(y_1) \rightarrow 1$ and $\pi^*(y_2) \rightarrow 0$, which is the deterministic optimal policy. The regularisation parameter $\tau$ can now actually be used to control how close to $\piref$ we are.

\subsection{Sampled Preferences}
\label{sec:sampled preferences}

So far we relied on the closed-form optimal policy from Eq.~\eqref{eq:pi-opt} to study \dpo and \ipo's stability, but this equation is not applicable to more complex settings where we only have access to sampled preference instead of $p^\star$.
We can still however find accurate approximations of the optimal policy by  choosing a parametrisation $\pi_\theta$ and optimize $\theta$ with an empirical loss over a dataset and iterative gradient-based updates.
We will use this approach to show two non-asymptotic examples where \dpo over-fits the dataset of preferences and ignore $\piref$: when one action $y$ wins against all others \dpo pushes $\pi_\theta(y)$ to 1 regardless of $\tau$, and conversely when one action $y$ never wins against the others \dpo pushes $\pi_\theta(y)$ to 0 again regardless of $\tau$.
In the same scenarios, \ipo does not converge to these degenerate solutions but instead remains close to $\piref$ based on the strength of the regularisation $\tau$.

For both scenarios we consider a discrete space $\mathcal{Y} = \{y_a, y_b, y_c\}$ with 3 actions, and select a dataset of pairs $\mathcal{D} = \{(y_{w,i}, y_{l,j})\}$. 
Given $\mathcal{D}$, we leverage the empirical losses from Eq.~\ref{eq:dpo-obj} and Eq.~\ref{eq:ipo-loss} to find \dpo's and \ipo's optimal policy.
We encode policies as $\pi_\theta (y_i) = \text{softmax}(\theta)_i$ using a vector $\theta \in \mathbb{R}^3$, and optimize them for $18000$ steps using Adam \citep{kingma2014adam} with learning rate $0.01$ and mini-batch size $9$. Mini-batches are constructed using uniform sampling with replacement from $\mathcal{D}$.
Both policies and losses are implemented using the \texttt{flax} python framework~\citep{jax2018github, flax2020github}, and the Adam implementation is from~\texttt{optax}~\citep{babuschkin2020deepmind}. For each set of hyper-parameters we repeat the experiment 10 times with different seeds, and report mean and 95\% confidence intervals. All experiments are executed on a modern cloud virtual machine with 4 cores and 32GB of ram.

\begin{figure}[t]
    \includegraphics[width=\columnwidth]{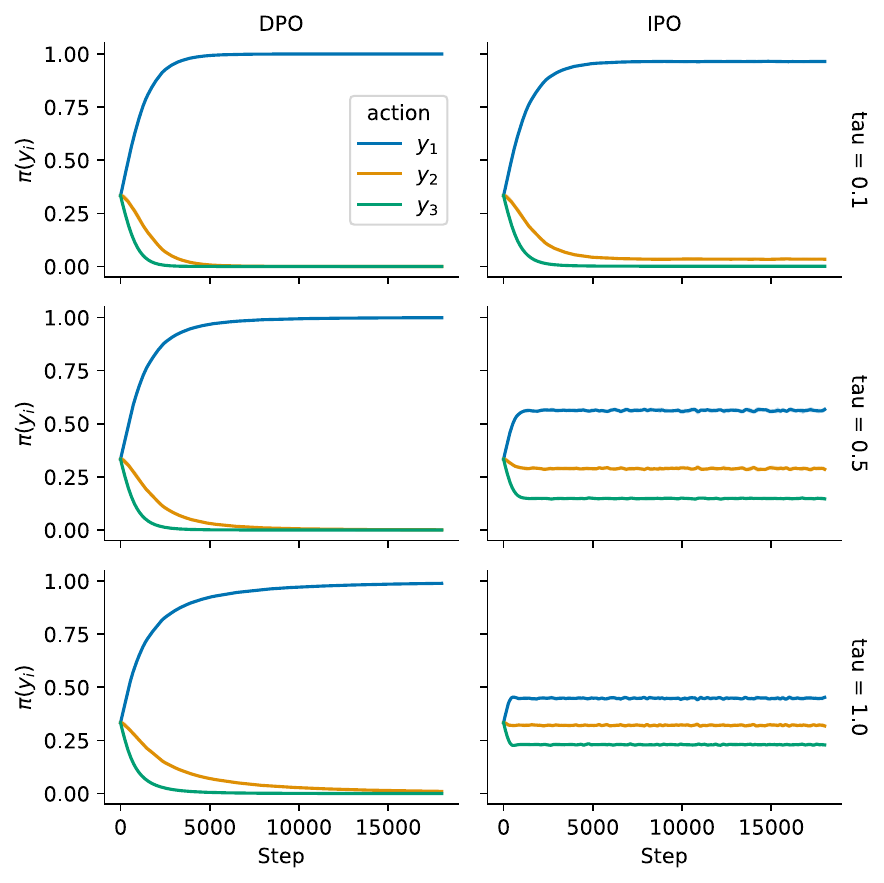}
    \vspace{-2\baselineskip}
    \caption{Comparison Between the Learning Curves of Action Probabilities of \ipo and \dpo for $\mathcal D_1$}
    \label{fig:exp_deterministic}
\end{figure}

\paragraph{IPO Avoids Greedy Policies}
For the first example we sample each unique action pair once to collect a dataset $\mathcal{D}$ containing 3 observed preferences.
Due to symmetries of pairwise preferences 
sampling only 3 preferences can results in only two outcomes (up to permutations of the actions):
\begin{align*}
    \mathcal{D}_1 = \{(y_a, y_b), (y_b, y_c), (y_a, y_c)\},\\
    \mathcal{D}_2 = \{(y_a, y_b), (y_b, y_c), (y_c, y_a)\},
\end{align*}
where we focus on $\mathcal{D}_1$, which represent a total ordering, rather than $\mathcal{D}_2$, which represent a cycle.
The outcome of the experiment is reported in Fig.~\ref{fig:exp_deterministic} in which, we report the learning curves for varying values of $\tau$. We observe that \dpo always converges to the deterministic policy for all values of $\tau$. In other word \dpo completely ignores the reference policy, no matter how strong is the regularisation term, and converges to the action which is preferred in the dataset. On the other hand, \ipo  prevent the policy from becoming greedy when the regularisation is strong.

\paragraph{\ipo Does not Exclude Actions}
In the first example \dpo converges to a deterministic policy because one action strictly dominates all others and the loss continues to push up its likelihood until it saturates.
The opposite effect happens for the logical opposite condition, i.e., when one action does not have at least a victory in the dataset \dpo will sets its probability to 0 regardless of $\tau$. While this is less disruptive than the first example (a single probability is perturbed whereas previously the whole policy was warped by an over-achieving action) it is also much more common in real-world data. In particular, whenever the action space is  large but the dataset small, some actions will necessarily be sampled rarely or only once, making it likely to never observe a victory. Especially because we do not have data on their performance $\pi$ should stick close to $\piref$ for safety, but \dpo's objective does not promote this.

\begin{figure}[t]
    \includegraphics[width=\columnwidth]{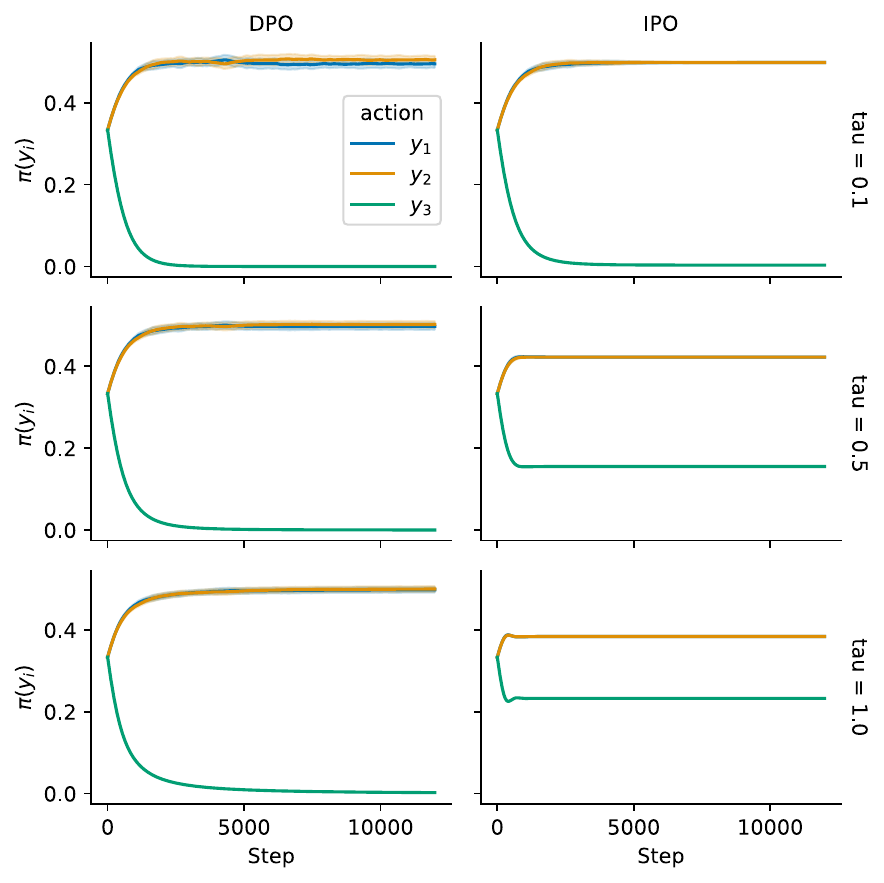}
    \vspace{-2\baselineskip}
    \caption{Comparison Between the Learning Curves of Action Probabilities of \ipo and \dpo for $\mathcal D_3$}
    \label{fig:exp_zero_prob}
\end{figure}

In the final example the dataset  consists of two observed preferences $\mathcal{D}_3 = \{(y_a, y_b), (y_b, y_a)\}$ and leave the pair $(y_a, y_c)$ completely unobserved. We compute solutions using Adam once again, and report the results in Fig.~\ref{fig:exp_zero_prob} for varying values of $\tau$. We observe again here that  \dpo  ignores the prior $\piref$ completely, no matter how strong we regularize the objective, whereas \ipo gradually decreases the probability of unobserved action with $\tau$.

\section{Conclusion and Future Work}
We presented a unified objective, called \sipo, for learning from preferences. It unifies \rlhf and \dpo methods. In addition, we introduced a particular case of \sipo, called \ipo, that allows to learn directly from preferences without a reward modelling stage and without relying on the Bradley-Terry modelisation assumption that assumes that pairwise preferences can be substituted with pointwise rewards. This is important because it allows to avoid the overfitting problem. This theoretical contribution is only useful in practice if an empirical sampled loss function can be derived. This is what we have done in Sec~\ref{sec:ipo} where we show that \ipo can be formulated as a root-finding problem from which an empirical sampled loss function can be derived. The \ipo loss function is simple, easy to implement and theoretically justified. Finally, in Sec.~\ref{sec:illustrative examples} and Sec.~\ref{sec:sampled preferences}, we provide illustrative examples where we highlight the instabilities of \dpo when the preferences are fully-known as well as when they are sampled. Those minimal experiments are sufficient to prove that \ipo is better suited to learn from sampled preferences than \dpo. Future works should scale those experiments to more complex settings such as training language models on human preferences data.

\clearpage

\bibliographystyle{plainnat}
\bibliography{main}

\begin{appendix}

\onecolumn
\section*{\centering APPENDICES}

\section{Proofs}

\subsection{Existence and uniqueness of the regularized argmaximum}
\label{appendices: argmaximum}
For completeness, we briefly recall the proof of existence and uniqueness of the argmaximum of the following regularized criterion that can also be found in the work of \citet{rafailov2023direct}:
\begin{align*}
\mathcal{L}_\tau(\delta) &= \mathbb{E}_{s\in\delta}[f(s)]-\tau \text{KL}(\delta \; || \; \eta),
\\
&= \sum_{s\in\mathcal{S}}\delta(s)f(s)-\tau \text{KL}(\delta \; || \; \eta),
\end{align*}
where $\mathcal{S}$ is a finite set, $f\in\mathbb{R}^\mathcal{S}$ a function mapping elements of $\mathcal{S}$ to real numbers, $\tau\in\mathbb{R}^*_+$ a strictly positive real number, $\delta\in\Delta_{\mathcal{S}}$ and $\eta\in\Delta_{\mathcal{S}}$ are discrete probability distributions over $\mathcal{S}$. In particular, we recall that a discrete probability distribution $\delta\in\Delta_{\mathcal{S}}$ can be identified as a positive real function $\delta\in\mathbb{R}_+^{\mathcal{S}}$ verifying:
\begin{equation*}
    \sum_{s\in\mathcal{S}}\delta(s) = 1.
\end{equation*}

Now, if we define the softmax probability $\delta^*\in\Delta_{\mathcal{S}}$ as:
\begin{equation*}
    \forall s\in\mathcal{S}, \delta^*(s) = \frac{\eta(s)\exp(\tau^{-1}f(s))}{\sum_{s'\in\mathcal{S}}\eta(s')\exp(\tau^{-1}f(s'))},
\end{equation*}

then, under the previous definitions, we have the following result:
\begin{equation*}
    \delta^*=\argmax_{\delta\in\Delta_{\mathcal{S}}}\mathcal{L}_\tau(\delta)
\end{equation*}

\begin{proof}
\begin{align*}
\frac{\mathcal{L}_\tau(\delta)}{\tau} &=  \sum_{s\in\mathcal{S}}\delta(s)\frac{f(s)}{\tau} - \text{KL}(\delta \; || \; \eta),
\\
&=\sum_{s\in\mathcal{S}}\delta(s)\frac{f(s)}{\tau}- \sum_{s\in\mathcal{S}}\delta(s)\log\big(\frac{\delta(s)}{\eta(s)}\big),
\\
&=\sum_{s\in\mathcal{S}}\delta(s)\big(\frac{f(s)}{\tau}- \log\big(\frac{\delta(s)}{\eta(s)}\big)\big) ,
\\
&=\sum_{s\in\mathcal{S}}\delta(s)\big(\log\big(\exp(\tau^{-1}f(s))\big)- \log\big(\frac{\delta(s)}{\eta(s)}\big)\big) ,
\\
&=\sum_{s\in\mathcal{S}}\delta(s)\big(\log\big(\frac{\eta(s)\exp(\tau^{-1}f(s))}{\delta(s)}\big)\big),
\\
&=\sum_{s\in\mathcal{S}}\delta(s)\big(\log\big(\frac{\eta(s)\exp(\tau^{-1}f(s))\frac{\sum_{s'\in\mathcal{S}}\eta(s')\exp(\tau^{-1}f(s'))}{\sum_{s'\in\mathcal{S}}\eta(s')\exp(\tau^{-1}f(s'))}}{\delta(s)}\big)\big),
\\
&=\sum_{s\in\mathcal{S}}\delta(s)\big(\log\big(\frac{\frac{\eta(s)\exp(\tau^{-1}f(s))}{\sum_{s'\in\mathcal{S}}\eta(s')\exp(\tau^{-1}f(s'))}}{\delta(s)}\big)\big) + \sum_{s\in\mathcal{S}}\delta(s)\log\big(\sum_{s'\in\mathcal{S}}\eta(s')\exp(\tau^{-1}f(s'))\big),
\\
&=\sum_{s\in\mathcal{S}}\delta(s)\big(\log\big(\frac{\delta^*(s)}{\delta(s)}\big)\big) + \log\big(\sum_{s'\in\mathcal{S}}\eta(s')\exp(\tau^{-1}f(s'))\big),
\\
&=-\text{KL}(\delta \; || \; \delta^*)+ \log\big(\sum_{s'\in\mathcal{S}}\eta(s')\exp(\tau^{-1}f(s'))\big).
\end{align*}

By definition of the KL, we now that $\delta^*=\argmax_{\delta\in\Delta_{\mathcal{S}}}\bigg[-\text{KL}(\delta \; || \; \delta^*)\bigg]$ and as:
\begin{equation*}
-\text{KL}(\delta \; || \; \delta^*) =\frac{\mathcal{L}_\tau(\delta)}{\tau}- \log\big(\sum_{s'\in\mathcal{S}}\eta(s')\exp(\tau^{-1}f(s'))\big)
\end{equation*}
where $\log\big(\sum_{s'\in\mathcal{S}}\eta(s')\exp(\tau^{-1}f(s'))\big)$ is a constant (does not depend on $\delta$) and $\tau$ a positive multiplicative term, then $-\text{KL}(\delta \; || \; \delta^*)$ and $\mathcal{L}_\tau(\delta)$ share the same argmaximum. This concludes the proof.
\end{proof}

\subsection{Non-uniqueness when $\texttt{Supp}(\pi(\cdot))\neq \texttt{Supp}(\mu)$:}
Notice that if we search for a solution where the support of $\pi$ is strictly larger than that of $\mu$ then there could be multiple solutions. Let us illustrate this case with a simple example. Consider a single state $x$ and 3 actions $y_1, y_2, y_3.$ The reference policy $\piref$ is uniform over $\{y_1, y_2, y_3\}$ and the policy $\mu$ assigns a probability $1/2$ to both $y_1$ and $y_2$ and $0$ probability to $y_3$.

Thus the loss is $L(\pi)=2\Big(\tau^{-1}\big(p^*(y_1\succ\mu)-p^*(y_2\succ\mu)\big)-\log\frac{\pi(y_1)}{\pi(y_2)}\Big)^2$. We deduce that any policy $\pi=(p,q,1-p-q)$ such that $\frac{p}{q}=e^{\tau^{-1}(p^*(y_1\succ\mu)-p^*(y_2\succ\mu))}$ is a global minimum of $L(\pi)$. 

In particular there are an infinity of solutions different from the optimal solution $\pi^*$. The problem comes from the fact that when the support of $\mu$ does not cover the whole action space there are not enough constraints to uniquely characterize $\pi^*$. Assuming that the supports of $\piref$ and $\mu$ coincide enables us to recover uniqueness of the solution, as proven in Theorem \ref{th:root finding}.

\section{Additional results}\label{sec:additional-results}

In this section, we show the equivalence of \dpo and \rlhf, regardless of whether the preference model $p^*$ corresponds to a Bradley-Terry model. Note that the assumption of the existence of a minimizer is to exclude cases where the loss is minimized by taking the rewards of certain actions to $+/- \infty$.

\begin{proposition}\label{prop:dpo-rlhf}
    Consider a preference model $p^*$ such that there exists a minimizer to the Bradley-Terry loss
    \begin{align*}
        \argmin_r \qquad  -\E_{\substack{x \sim \rho \\ y \sim \mu(\cdot|x) \\ y' \sim \mu(\cdot|x) }}
        [p^*(y \succ y'|x) \log \sigma(r(x, y) - r(x, y'))] \, .
    \end{align*}
    Then, the optimal policy for the \dpo objective in Equation~\eqref{eq:dpo-obj} and for the \rlhf objective in Equation~\eqref{eq:rlhf-obj} with reward model given as the minimizer to the Bradley-Terry loss above are identical, regardless of whether or not $p^*$ corresponds to a Bradley-Terry preference model.
\end{proposition}

\begin{proof}
    Recall that the optimal policy $\pi^*_r$ for a given reward function $r$ for the objective in Equation~\eqref{eq:rlhf-obj} is given by $\pi^*_r(y|x) \propto \piref(y|x) \exp(\tau^{-1} r(x, y))$. It therefore follows that
    \begin{align*}
        & - \E_{\substack{x \sim \rho \\ y, y' \sim \mu(\cdot|x) }}
        [p(y \succ y'|x) \log \sigma(r(x, y) - r(x, y'))] \\
        = & - \E_{\substack{x \sim \rho \\ y, y' \sim \mu(\cdot|x) }}
        \Big[
            p(y \succ y'|x) \log \sigma\Big(
                \tau \log\bigg( \frac{\pi^*_r(y|x)}{\pi^*_r(y'|x)}\bigg ) - 
                \tau \log\bigg( \frac{\piref(y|x)}{\piref(y'|x)}\bigg ) 
            \Big)
        \Big] \, .
    \end{align*}
    In words, the value of the Bradley-Terry reward objective for $r$ is the value of the \dpo objective for $\pi^*_r$. We recall also that the map $r \mapsto \pi^*_r$ is surjective.
    
    Now, suppose $r$ is optimal for the Bradley-Terry reward objective, meaning that $\pi^*_r$ is optimal for the \rlhf objective. If $\pi^*_r$ is not optimal for the \dpo objective, then there exists another policy $\pi'$ that obtains a strictly lower value for the \dpo loss. But then there exists a reward function $r'$ such that $\pi' = \pi^*_{r'}$, such as $r'(x, y) = \tau \log(\pi'(y|x)/\piref(y|x))$, and this $r'$ therefore obtains a lower Bradley-Terry loss than $r$, a contradiction.
    
    Similarly, if $\pi^*$ is optimal for the \dpo objective, the corresponding reward function $r(x, y) = \tau \log(\pi^*(y|x) / \piref(y|x))$ must be optimal for the Bradley-Terry reward loss. The corresponding optimizer for the \rlhf objective is then given by $\pi(y|x) \propto \piref(y|x) \exp(\tau^{-1} \tau \log(\pi^*(y|x)/ \piref(y|x))) = \pi^*(y|x)$, as required.
\end{proof}

\end{appendix}

\end{document}